\documentclass[final,5p,times,twocolumn]{elsarticle}

\usepackage{amssymb}
\usepackage{amsmath}
\usepackage{graphicx}%
\usepackage{multirow}%
\usepackage{amsmath,amsfonts}%
\usepackage{amsthm}%
\usepackage{mathrsfs}%
\usepackage[title]{appendix}%
\usepackage{xcolor}%
\usepackage{textcomp}%
\usepackage{manyfoot}%
\usepackage{booktabs}%
\usepackage{algorithm}%
\usepackage{algorithmicx}%
\usepackage{algpseudocode}%
\usepackage{listings}%
\usepackage{makecell}%
\usepackage{hyperref}

\newif\ifreview
\reviewfalse   
\newcommand{\rev}[1]{\ifreview\textcolor{red}{#1}\else#1\fi}

\newtheorem{theorem}{Theorem}
\newtheorem{proposition}{Proposition}

\newtheorem{assumption}{Assumption}




\begin{document}

\begin{frontmatter}

\title{Enhancing Visual Feature Attribution via Weighted Integrated Gradients}

\author[a]{Tran Duc Tuan Kien}
\author[a]{Nguyen Trong Tam}
\author[a]{Nguyen Hoang Son}
\author[a]{Khoat Than}
\author[a]{Duc Anh Nguyen\corref{cor1}}

\affiliation[a]{organization={Hanoi University of Science and Technology},
            addressline={No 1 Dai Co Viet}, 
            city={Hanoi},
            postcode={100000}, 
            state={Hanoi},
            country={Vietnam}}
\cortext[cor1]{Corresponding author. Email: anhnd@soict.hust.edu.vn. The paper is under consideration at Pattern Recognition Letters.}

\begin{abstract}
Integrated Gradients (IG) is a widely used attribution method in explainable AI, particularly in computer vision applications where reliable feature attribution is essential. A key limitation of IG is its sensitivity to the choice of baseline (reference) images. Multi-baseline extensions such as Expected Gradients (EG) assume uniform weighting over baselines, implicitly treating all baseline images as equally informative. In high-dimensional vision models, this assumption often leads to noisy or unstable explanations. This paper proposes {Weighted Integrated Gradients} (WG), a principled approach that evaluates and weights baselines to enhance attribution reliability. WG introduces an unsupervised criterion for baseline suitability, enabling adaptive selection and weighting of baselines on a per-input basis. \rev{The method preserves the core axiomatic properties of IG in a generalized weighted-baseline form. Under an expected, proxy-based fitness--relevance monotonicity assumption, WG provides a probabilistic justification for assigning larger weights to more informative baselines. Experiments on commonly used image datasets and models show that WG improves over EG under our protocol, with up to 36\% gains across evaluated convolutional and Transformer architectures. These gains come with additional fitness-evaluation cost, so WG should be viewed as an attribution-fidelity trade-off rather than a faster alternative to EG.} By moving beyond the assumption that all baselines contribute equally, Weighted Integrated Gradients offers a clearer and more reliable approach to explaining computer-vision models, improving both understanding and practical usability in explainable AI.
\end{abstract}


\end{frontmatter}

\section{Introduction}

Explainable AI (XAI) is increasingly important as deep neural networks, especially computer vision tasks such as classification, detection, and segmentation. In these applications, understanding model behavior is essential for trust, safety, and diagnostics~\cite{dwivedi2024efficient,koenen2024toward}. A central direction in XAI is {feature attribution}, which quantifies how individual pixels or regions contribute to a model's output. Among attribution methods, {Integrated Gradients} (IG)~\cite{Sundararajan16} is a key technique, built on path-based attribution and related to Shapley values~\cite{sundararajan2020many}. IG integrates gradients along a straight-line path from a baseline to the input to derive feature importance.

While IG satisfies several desirable axioms, it faces practical challenges. Prior work highlights three issues for path-based methods: baseline choice, path construction, and counterintuitive attributions~\cite{Akhtar23}. Of these, {baseline selection} is the most critical in vision, where no universally "neutral" reference image exists. Common baselines such as black, white, blurred, or random images often yield inconsistent explanations, and the theoretical link between baseline choice and attribution behavior remains poorly understood \cite{sturmfels2020visualizing}. 

\rev{Integrated Gradients is sensitive to the choice of baseline, while Expected Gradients reduces this sensitivity by averaging over multiple baselines~\cite{Erion21}. However, Expected Gradients assigns equal weights to all baselines, although different baselines may not be equally informative for a specific input; this motivates an input-dependent baseline-weighting approach.} In high-dimensional vision models, baselines vary greatly in their relevance. When many are low quality, uniform averaging can distort explanations by allowing poor baselines to dominate or dilute the final attribution.

We propose Weighted Integrated Gradients (WG), \rev{a baseline-aware extension of IG that assigns input-dependent weights to different reference baselines. The key contribution is not to introduce a new attribution path, but to address the uniform-baseline assumption in EG by estimating how suitable each baseline is for explaining a specific input.} Our key insight is that baseline quality depends jointly on both the input and the baseline; hence each baseline should contribute proportionally to its explanatory relevance. Unlike EG, WG assigns \emph{data-driven weights} that emphasize baselines producing more faithful explanations.

We define an unsupervised fitness function that evaluates baseline suitability solely from model behavior, requiring no annotation or domain knowledge. \rev{The fitness score is estimated by a binary-search procedure that reduces the number of perturbation steps per baseline; nonetheless, WG still requires additional model evaluations compared with EG. We therefore present WG as an attribution-fidelity trade-off, intended for settings where improved explanation quality justifies the additional runtime.} These scores are then normalized into weights used to aggregate IG attributions. Experiments on standard vision benchmarks show that WG improves attribution accuracy, fidelity, and stability over existing multi-baseline methods.

Beyond improved interpretability, the framework naturally identifies and down-weights low-quality baselines, ensuring reliable attributions without manual baseline curation. This adaptive selection makes the method well suited for modern vision models requiring robust explanation tools. Experiments show that WG boosts attribution quality relatively by up to 36\% across diverse models. WG preserves IG's theoretical foundations while adding a principled, adaptive, data-driven weighting mechanism, resulting in an accurate attribution method for deep learning systems.

\rev{Our contribution is threefold. First, we formulate baseline selection in IG as an input-dependent weighting problem and propose a perturbation-based fitness score for estimating baseline usefulness. Second, we provide an assumption-based analysis showing that, under expected proxy fitness--relevance monotonicity, fitness-weighted aggregation can increase expected relevance mass compared with uniform baseline averaging. Third, we empirically evaluate WG across convolutional and Transformer backbones, reporting both its attribution-fidelity gains and its computational overhead.}

\section{Related work}

Explainable AI (XAI) methods are often grouped into perturbation-based, backpropagation-based, and attention-based approaches. Perturbation methods estimate feature importance by masking or altering inputs but are computationally expensive for high-resolution models~\cite{Li16}. Backpropagation-based techniques propagate relevance using gradients or attribution rules and offer a far more efficient alternative~\cite{smilkov2017smoothgrad,yang2023idgi}. Attention-based methods highlight influential tokens via transformer attention, though their interpretability remains debated~\cite{Vig19}. Among these, gradient-based backpropagation strikes the best balance between efficiency and interpretability and forms the foundation of our approach.

\emph{Integrated Gradients} (IG)~\cite{Sundararajan16} attributes importance by integrating gradients from a baseline to the input, but its effectiveness depends heavily on the chosen baseline~\cite{Akhtar23}, which is particularly problematic in vision tasks lacking a neutral reference image. To address this, prior work has explored improved baseline strategies and multi-baseline formulations, such as learned or sampled baselines~\cite{Erion21}, as well as path-based refinements that better align with perceptual structure~\cite{Xu20}. While some methods introduce weighting schemes for attribution, none explicitly focus on weighting IG baselines. More broadly, fitness-based weighting connects to recent work on uncertainty-aware feature scoring~\cite{wan2026uncertainty}, which designs global weighting factors via fuzzy multi-granularity information measures, and to decoupled dual-branch architectures~\cite{wang2026twin} that separate scoring from downstream processing.

We fill this gap with \emph{Weighted Integrated Gradients} (WG), a framework that assigns input-dependent baseline weights using an unsupervised fitness measure. Unlike EG's uniform averaging or SIG's sampling-based proportionality, WG selectively emphasizes baselines that yield faithful local explanations, improving reliability and stability. 

\section{Background}

In this section, we briefly review the key concepts of Integrated Gradients and Expected Gradients, and then highlight the limitations of the latter. Reformulating Integrated Gradients as a gradient path integral and Expected Gradients as a uniform expectation over baselines reveals a central drawback of EG: its assumption that all baselines contribute equally, regardless of their relevance to a specific input. This oversimplified weighting scheme leads to suboptimal and unstable attributions. These limitations motivate our proposed method, which introduces refined, input-aware baseline weighting for more accurate and reliable explanations.

\subsection{Integrated Gradients}
Integrated Gradients (IG) \cite{Sundararajan16} computes feature attributions by accumulating gradients along a path from a baseline $x'$ to the input $x$. Inspired by game-theoretic foundations, IG begins at a reference point, ideally a neutral state carrying minimal information, and traces a continuous trajectory toward $x$, capturing how features contribute to the model's prediction.

Formally, for a classifier, a class $c$, and an input $x$, let $f_c(x)$ denote the model's score for class $c$. The attribution for the $i$th feature is defined as:

\begin{equation}
    \text{IG}_i(x, x') := (x_i - x'_i) \times \int_0^1 \frac{\partial f(x' + \alpha(x - x'))}{\partial x_i} \, d\alpha,
\end{equation}
where $\frac{\partial f(x)}{\partial x_i}$ is the gradient of $f$ along the $i$th feature and $\alpha$ parameterizes the interpolated coefficient of points that lie between input and the reference point $x'$ ($\alpha \in [0,1]$). This integral captures each feature's contribution by averaging its gradient's influence across the path.

Empirical evidence underscores that baseline choice profoundly impacts explanation quality, yet no rigorous theory defines an optimal \( x' \) for arbitrary inputs. A baseline effective for one image may falter for another, necessitating laborious, image-specific tuning that remains suboptimal and impractical. This variability exposes a gap in path-based XAI: the lack of a principled, adaptive strategy for baseline selection.

\subsection{Expected Gradient}

To address the baseline selection issue in Integrated Gradients (IG), Expected Gradients (EG)~\cite{Erion21} averages IG over multiple baselines instead of relying on a single one. By uniformly sampling baselines from a pre-selected set $D' = \{x^{(k)}\}_{k=1}^n$, EG approximates the expected attribution as $\mathrm{EG}_i(x) = \int_{x'} \mathrm{IG}_i(x, x') p_D(x') \, dx' \approx \frac{1}{n}\sum_{k=1}^{n}\mathrm{IG}_i(x, x^{(k)})$.

This approach captures diverse perspectives from multiple reference points, producing more stable explanations. However, EG assigns uniform weights to all baselines, treating them as equally informative even though some align well with decision boundaries while others introduce noise. As a result, uniform averaging can dilute useful signals and distort the model's rationale.

To address this, we propose Weighted Integrated Gradients (WG), which assigns input-specific weights based on each baseline's ability to yield accurate, informative gradients. By emphasizing high-fitness baselines, WG produces more precise and interpretable attributions, improving the fidelity of gradient-based explanations.

\section{Proposed Method}

In this section, we propose Weighted Integrated Gradients (WG), a novel advancement that redefines baseline selection in Integrated Gradients (IG) through a perturbation-based fitness measure. WG departs from the limitations of prior methods by assigning weights to baselines according to their explanatory fidelity, thereby enhancing the precision and robustness of feature attributions.

Consider a differentiable model $f: \mathbb{R}^d \rightarrow \mathbb{R}$, an input $x \in \mathbb{R}^d$, and a set of baselines $D' = \{x^{(k)}\}_{k=1}^n$. The WG attribution for feature $(i)$ is defined as:

\begin{align}
    \text{WG}_i (x) &= \sum_{k=1}^{n} w_k \cdot \text{IG}_i(x, x^{(k)}), \label{eq:wg1} \\
    &= \sum_{k=1}^{n} w_k (x_i - x^{(k)}_i) \int_0^1 \frac{\partial f(x^{(k)} + \alpha (x - x^{(k)}))}{\partial x_i} d\alpha \label{eq:wg2}
\end{align}
where $w_k$ represents the weight of the $(k)$-th baseline $x^{(k)}$, derived from its fitness, and the integral computes the IG attribution for each baseline. By modulating the influence of each baseline via $w_k$, WG ensures that more informative reference points dominate the aggregation, yielding a final explanation that better reflects the model's decision-making process. In the following, we describe a strategy for calculating $w_k$.

\subsection{Fitness Measure Based on Perturbation}

Perturbation offers a natural and intuitive method for evaluating individual elements in complex systems. In explainable AI, particularly for baseline selection in Integrated Gradients (IG), it helps assess each baseline's impact by modifying or removing input features and observing the resulting output changes.

\rev{This method does not require manual annotations and relies on model-output changes to estimate baseline usefulness. However, it should be understood as a proxy criterion for attribution fidelity rather than a direct measurement of the true causal relevance of a baseline.} To quantify the quality of a baseline for a given input, we propose a fitness function \( D_\alpha \), defined as the minimum number of features that must be masked to reduce the model's output to \( \alpha\% \) of its original score:
\begin{align}
D_{\alpha} = \min \left\{ |S| : S \subseteq F,\ f(x \circ M(S)) \leq \alpha \cdot f(x) \right\} \label{eq:dalpha}
\end{align}
where \( M(S) \) masks features in \( S \), and \( x \circ M(S) \) denotes the masked input. A smaller \( D_\alpha \) implies the baseline effectively highlights influential features, since fewer need to be masked to impact the output significantly.

In Weighted Integrated Gradients (WG), we assign weights based on this fitness:
\begin{align}
w_k \propto \frac{1}{D_\alpha}
\end{align}
ensuring that baselines with higher explanatory power (lower \( D_\alpha \)) contribute more to the final attribution. This approach is both principled and unsupervised, aligning attribution fidelity with model sensitivity.

To compute $D_\alpha$, the minimal number of features to mask to reduce the model's output score to $\alpha\%$ of its original value, a naive exhaustive search would incur a prohibitive $O(n)$ time complexity, where $n$ is the number of features. Leveraging the monotonicity of score reduction upon feature removal, we propose Algorithm \ref{alg:da}, an efficient binary search strategy, achieving a time complexity of $O(\log n)$ per baseline. The algorithm iteratively adjusts the proportion of features masked—guided by their importance scores (e.g., IG attributions)—until the target score is approximated within a tolerance $\epsilon$.

\subsection{Algorithm for Efficient Fitness Calculation}
\renewcommand{\alglinenumber}[1]{\footnotesize\text{#1}}
\begin{algorithm}[H]
\fontsize{8}{8} \selectfont
\caption{Compute\_D\_alpha}
\begin{algorithmic}[1]
\Require $x$: input data (e.g., image)
\Require $attr$: feature importance scores (e.g., IG attributions)
\Require $f$: model function (classifier)
\Require $\alpha$: target score reduction ratio $(0 < \alpha < 1)$
\Require $neutral$: neutral value for masked features
\Require $\varepsilon$: convergence threshold
\Ensure $d_\alpha$: minimal number of removal pixels
\Statex \text{0:} $\textit{target\_score} \gets \alpha \cdot f(x)$;\ $\textit{low} \gets 0$;\ $\textit{high} \gets \text{count}(x)$
\Statex \text{1:} \textbf{while} $low \leq high$ \textbf{do}
\Statex \text{2:} \quad $mid \gets \lfloor (low + high) / 2 \rfloor$
\Statex \text{3:} \quad $mask \gets \text{FindMask}(attr, mid)$
\Statex \text{4:} \quad $x' \gets x \cdot (1 - mask) + neutral \cdot mask$
\Statex \text{5:} \quad \textbf{if} $|f(x') - target\_score| < \varepsilon$ \textbf{then}
\Statex \text{6:} \quad\quad \textbf{return} $mid$
\Statex \text{7:} \quad \textbf{else if} $f(x') < target\_score$ \textbf{then}
\Statex \text{8:} \quad\quad $low \gets mid + 1$
\Statex \text{9:} \quad \textbf{else}
\Statex \text{10:} \quad\quad $high \gets mid - 1$
\Statex \text{11:} \quad \textbf{end if}
\Statex \text{12:} \textbf{end while}
\Statex \text{13:} \textbf{return} $mid$
\end{algorithmic}
\label{alg:da}
\end{algorithm}

\section{Theoretical Analysis}
\label{sec:theory}
Since WG is defined as a non-negative weighted sum of Integrated Gradients (IG) across multiple baselines, and IG satisfies \emph{Implementation Invariance} and \emph{Effectiveness}~\cite{Sundararajan16}, WG naturally inherits these two properties. 
Regarding completeness, IG satisfies the standard axiom with respect to a \emph{single} baseline, whereas WG satisfies an analogous \emph{generalized completeness} property: the total WG attribution equals the model output at $x$ minus the \emph{weighted expected baseline output}. 

Beyond these axiomatic considerations, we further study WG from a probabilistic viewpoint. 
Each IG attribution vector is interpreted as a probability distribution over features, and we assume that baselines with better fitness scores $D_\alpha$ tend to place more probability mass on truly relevant features. 
Under this mild assumption, we show that WG increases the expected relevance of sampled baselines and provides tighter finite-sample probabilistic guarantees than equal-weight averaging (EG). 
For detailed statements and proofs, please refer to~\ref{appendix:proof}.

\rev{The analysis below is an expected, proxy-based justification rather than a per-instance correctness guarantee. Since the truly relevant feature set is generally unobservable for real vision models, both the assumption and its empirical verification rely on proxy relevance sets, such as segmentation masks.}

\subsection{Probabilistic model}

For each baseline $x^{(k)}$, we define the normalized positive IG profile as $p^{(k)}_j=\dfrac{\big[ IG_j(x,x^{(k)}) \big]_{+}}{\sum_{\ell=1}^d \big[ IG_\ell(x,x^{(k)}) \big]_{+}}$ for $j=1,\dots,d$, where $[a]_{+}:=\max(a,0)$ denotes the positive part. By construction, $\sum_{j=1}^d p^{(k)}_j=1$.

Interpret $p^{(k)}$ as a probability distribution on feature indices. Let $R(x)\subseteq\{1,\dots,q\}$ be the (unknown) set of truly relevant features. Define the relevance quality $Q_k(x)=\sum_{j\in R(x)} p^{(k)}_j=\Pr_{J\sim p^{(k)}}(J\in R(x))$. Let $u_k=1/n$ be the EG weights and $w_k=\dfrac{D_\alpha(x^{(k)})^{-1}}{\sum_{j=1}^n D_\alpha(x^{(j)})^{-1}}$ be the WG weights derived from the fitness scores $D_\alpha(x^{(k)})$.

\begin{assumption}[\rev{Expected proxy} fitness--relevance monotonicity]
\label{ass:fitness_relevance}
For each input $x$, order baselines so that: $D_\alpha(x^{(1)}) \le D_\alpha(x^{(2)}) \le \cdots \le D_\alpha(x^{(n)}),$
and let $k(x,r)$ denote the baseline with rank $r$ for $x$. Assume that better fitness implies higher relevance \emph{in expectation over the data distribution} $\mathcal{D}$: for any pair of ranks $r < r'$ in $\{1,\dots,n\}$ : $
\mathbb{E}_{x\sim\mathcal{D}}\!\left[Q_{k(x,r)}(x)\right]
\;\ge\;
\mathbb{E}_{x\sim\mathcal{D}}\!\left[Q_{k(x,r')}(x)\right]$.
\end{assumption}

\rev{This expectation-level assumption is directly testable using proxy relevance sets and is empirically verified on all seven backbones (\ref{appendix:assumption_verification}); aggregate-level support does not imply that monotonicity holds for every individual input.}

\subsection{WG improves expected relevance}

Define the rank-marginal weights $\bar{W}_r := \mathbb{E}_{x\sim\mathcal{D}}\!\left[w_{k(x,r)}(x)\right]$ and the rank-marginal WG profile $p^{\mathrm{WG}^{\star}}_j(x) := \sum_{r=1}^n \bar{W}_r\, p^{(k(x,r))}_j$, with relevance probability $q_{\mathrm{WG}^{\star}}(x) := \sum_{j\in R(x)} p^{\mathrm{WG}^{\star}}_j(x)$.

\begin{proposition}[\rev{Expected relevance advantage under proxy monotonicity}]
\label{prop:wg_improves_relevance}
Under Assumption~\ref{ass:fitness_relevance}, we have:
\[
\mathbb{E}_{x}\!\left[q_{\mathrm{WG}^{\star}}(x)\right]
\;\ge\;
\mathbb{E}_{x}\!\left[q_{\mathrm{EG}}(x)\right],
\]
with strict inequality whenever $\bar{Q}_r := \mathbb{E}_x[Q_{k(x,r)}(x)]$ is non-constant in $r$.
\end{proposition}

Thus, aggregating baselines with rank-marginal WG weights yields, on average over the data distribution, higher mass on truly relevant features compared to EG. The gap between actual WG and its rank-marginal version is the per-rank covariance $\sum_r \mathrm{Cov}_x(w_{k(x,r)}(x), Q_{k(x,r)}(x))$, which is empirically favourable to WG in practice (Table~\ref{tb:auccompare}). (See~\ref{appendix:proof} for the proof.)

\subsection{Finite-sample advantage of WG}

Let the aggregated WG attribution have normalized profile $p^{\mathrm{WG}}$, and define its relevance probability as $q_{\mathrm{WG}}(x)=\sum_{j\in R(x)} p^{\mathrm{WG}}_j=\Pr_{J\sim p^{\mathrm{WG}}}(J\in R(x))$. Define $q_{\mathrm{EG}}(x)$ analogously from $p^{\mathrm{EG}}$. Sampling $m$ features independently from $p^{\mathrm{WG}}$ yields the empirical relevance fraction $\hat{q}_{\mathrm{WG}}(x,m)=\frac{1}{m}\sum_{t=1}^m \mathbf{1}\{J_t^{\mathrm{WG}}\in R(x)\}$. Let $\mu_{\mathrm{EG}} := \mathbb{E}_{X\sim\mathcal{D}}[q_{\mathrm{EG}}(X)]$ denote the average EG relevance.

\begin{theorem}[\rev{Conditional finite-sample relevance statement for WG}]
\label{thm:wg_probabilistic_rate}
Given an input $x$ with margin $\delta(x) := q_{\mathrm{WG}}(x) - \mu_{\mathrm{EG}} > 0$. For any $m\ge 1$ and $\eta\in(0,1)$,
\[
\Pr\!\left(\hat{q}_{\mathrm{WG}}(x,m) \le \mu_{\mathrm{EG}}\right) \le \exp(-2m\,\delta(x)^2).
\]
Equivalently, $m \ge \frac{1}{2\delta(x)^2}\log(1/\eta)$ suffices to guarantee $\Pr(\hat{q}_{\mathrm{WG}}(x,m) > \mu_{\mathrm{EG}}) \ge 1-\eta$.
\end{theorem}

\rev{Theorem~\ref{thm:wg_probabilistic_rate} is conditional on the event that WG attains a positive relevance margin over the average EG relevance, and should not be read as a universal per-instance guarantee; when such a margin exists, the empirical relevance fraction concentrates above the EG average at an exponential rate.} Proposition~\ref{prop:wg_improves_relevance} guarantees $\mathbb{E}_X[q_{\mathrm{WG}^{\star}}(X)] \ge \mu_{\mathrm{EG}}$, so the set of inputs satisfying $\delta(x) > 0$ carries non-trivial mass under $\mathcal{D}$ whenever actual WG retains the aggregate advantage of its rank-marginal counterpart, which Table~\ref{tb:auccompare} empirically confirms.

Under this simple probabilistic relevance model, WG improves both the
\emph{expected} relevance of sampled baselines and the \emph{finite-sample}
relevance probability of the aggregated attribution.  
In contrast, EG treats all baselines equally and therefore cannot exploit
variation in baseline quality, whereas WG systematically amplifies
high-fitness, high-relevance baselines.

\section{Experiments}
In this section, we present a comprehensive evaluation of the proposed \text{Weighted Integrated Gradients (WG)} method, benchmarking it against established explainability technique Expected Gradients (EG). WG augments Integrated Gradients with a \emph{fitness-weighted} baseline aggregation strategy that directly addresses the challenge of baseline selection. \rev{The aim of our experiments is to compare WG with EG under commonly used attribution metrics and to examine whether the proposed fitness score provides useful baseline weights. We report both attribution-quality metrics and computational cost.}\footnote{Our implementation is available at \url{https://github.com/KienTranDSAI/Weighted-Integrated-Gradients}}

\subsection{Experimental Setup}

\textbf{Baseline:} \\
We evaluate Weighted Integrated Gradients (WG) against prominent XAI methods, with a primary focus on Expected Gradients (EG) \cite{Erion21}, which already showed the advantage over existing explanation method \cite{klein2024navigating}. WG enhances this by weighting baselines via a fitness score to improve attribution fidelity. The baseline library $D'$ consists of four deterministic references, \texttt{black}, \texttt{white}, \texttt{median} (per-channel median of $x$), and \texttt{bg\_mean} (mean RGB of the four $10\%\times 10\%$ corner patches of $x$), plus a \texttt{random} component drawn pixel-wise i.i.d.\ from $\mathcal{U}[x_{\min}, x_{\max}]$, resampled when more baselines are needed. By default we resample twice, giving $n=6$ baselines.

\textbf{Models and Datasets:} \\
Experiments are conducted using pretrained PyTorch models across seven architectures: DenseNet121, MobileNetV2, ResNet50, ResNet101, VGG16, VGG19, and ViT-B/16. We evaluate on ImageNet and COCO, following prior works \cite{Sundararajan16}. To ensure interpretability-focused analysis, we select samples where the model prediction is correct, enabling a fair comparison of explanation quality across methods.

\vspace{5pt}
\textbf{Evaluation metrics:} 

\textbf{AUC of Deletion}. The AUC of Deletion (Del) measures explanation quality by evaluating how rapidly a model's confidence drops when important features (ranked by an attribution method $\phi(x)$) are progressively removed. For an input $x$ and class $c$, a perturbed version of the input is created by masking the top-$p\%$ features in $\phi(x)$ using a mask $M_p(\phi(x))$, and the model's retained confidence is $\text{Deletion Score}(x,\phi(x),p)=f_c\!\left(x \circ M_p(\phi(x))\right)$. The AUC of Deletion is defined by aggregating scores over all perturbation levels ($p\in[0,1]$): $\text{AUC of Deletion}=\int_0^1 \text{Deletion Score}(x,\phi(x),p)\,dp \approx \frac{1}{N}\sum_{i=1}^{N} f_c\!\left(x \circ M_{p_i}(\phi(x))\right)$ with $p_i = i/N$.

A lower AUC indicates better explanations, as important features cause sharper confidence drops when removed. This metric is threshold-free, robust to noise, and aligns with the principle that ideal attributions should most disrupt the prediction when key features are removed. It serves as a core evaluation tool in our analysis of WG's effectiveness.

\begin{table}[tp]
\centering
\fontsize{7}{7} \selectfont
\renewcommand{\arraystretch}{1.1}
\begin{tabular}{llcccc}
\hline
\textbf{Metrics} & \textbf{Model} & \textbf{EG} & \textbf{WG (Ours)} & \textbf{Rel. I (\%)} & \textbf{$p$-value} \\
\hline
\multirow{7}{*}{\makecell[l]{Del  ($\downarrow$)}}
  & DenseNet121 & 0.084 $\pm$ 0.095 & \textbf{0.064 $\pm$ 0.090} & 24.03 & 0.000 \\
\cmidrule{2-6}
  & MobileNetV2 & 0.021 $\pm$ 0.033 & \textbf{0.014 $\pm$ 0.024} & 29.57 & 0.011 \\
\cmidrule{2-6}
  & ResNet50    & 0.044 $\pm$ 0.068 & \textbf{0.029 $\pm$ 0.051} & 34.35 & 0.002 \\
\cmidrule{2-6}
  & ResNet101   & 0.057 $\pm$ 0.082 & \textbf{0.037 $\pm$ 0.072} & 36.01 & 0.000 \\
\cmidrule{2-6}
  & VGG16       & 0.035 $\pm$ 0.059 & \textbf{0.027 $\pm$ 0.053} & 23.25 & 0.002 \\
\cmidrule{2-6}
  & VGG19       & 0.029 $\pm$ 0.048 & \textbf{0.022 $\pm$ 0.037} & 24.63 & 0.018 \\
\cmidrule{2-6}
  & ViT-B/16    & 0.132 $\pm$ 0.160 & \textbf{0.106 $\pm$ 0.134} & 19.36 & 0.024 \\
\hline
\multirow{7}{*}{\makecell[l]{OL ($\uparrow$)}}
  & DenseNet121 & 0.418 $\pm$ 0.175 & \textbf{0.462 $\pm$ 0.177} & 10.59 & 0.001 \\
\cmidrule{2-6}
  & MobileNetV2 & 0.414 $\pm$ 0.194 & \textbf{0.459 $\pm$ 0.199} & 10.82 & 0.001 \\
\cmidrule{2-6}
  & ResNet50    & 0.390 $\pm$ 0.189 & \textbf{0.448 $\pm$ 0.186} & 14.85 & 0.000 \\
\cmidrule{2-6}
  & ResNet101   & 0.391 $\pm$ 0.188 & \textbf{0.463 $\pm$ 0.187} & 18.55 & 0.000 \\
\cmidrule{2-6}
  & VGG16       & 0.423 $\pm$ 0.180 & \textbf{0.488 $\pm$ 0.180} & 15.33 & 0.000 \\
\cmidrule{2-6}
  & VGG19       & 0.427 $\pm$ 0.181 & \textbf{0.459 $\pm$ 0.182} &  7.38 & 0.028 \\
\cmidrule{2-6}
  & ViT-B/16    & 0.387 $\pm$ 0.163 & \textbf{0.409 $\pm$ 0.171} &  5.69 & 0.046 \\
\hline
\end{tabular}
\caption{Comparison of Expected Gradients and Weighted Integrated Gradients performance across Models. For AUC of Deletion (Del), smaller is better; for AUC of Overlap (OL), higher is better. $p$-values from paired $t$-test EG vs WG.}
\label{tb:auccompare}
\end{table}

\textbf{AUC of Overlap}. While AUC of Deletion measures the effect of removing important features, it does not assess whether attributions align with human-recognizable regions. To address this, we introduce the \textit{AUC of Overlap} (OL), a supervised metric evaluating how well attribution maps match ground-truth segments. Given a segmentation mask $S$ of size $|S|$, let $T_p$ be the top $p|S|$ pixels of $\phi(x)$. The overlap at proportion $p$ is $\text{Overlap}(p)=\frac{|T_p \cap S|}{p|S|}$. The metric aggregates overlap across $p\in(0,1]$: $\text{AUC of Overlap}=\int_0^1 \text{Overlap}(p)\,dp \approx \frac{1}{N}\sum_{i=1}^N \text{Overlap}\!\left(\frac{i}{N}\right)$.

A higher AUC indicates stronger spatial alignment between the explanation and the ground-truth object, making this metric essential for evaluating interpretability in vision tasks. \rev{Overlap AUC has important limitations: it requires ground-truth segmentation masks; it measures spatial agreement with annotated object regions rather than direct causal faithfulness; and it is computed on correctly classified samples, which may introduce selection bias since misclassified examples are excluded. It should therefore be interpreted together with Deletion AUC, not as a standalone measure.}

\textbf{Fitness Function:}
 To compute the fitness measure \( D_{\alpha} \), we set the parameter \( \alpha = 0.5 \), corresponding to a 50\% reduction in model confidence--a threshold with deep scientific relevance. In fields like toxicology, the 50\% threshold (e.g., LC50 or LD50) is a standard benchmark for assessing median effects, reflecting a critical inflection point. In explainable AI, \( \alpha = 0.5 \) identifies the feature set whose removal halves the model's score, capturing a pivotal transition where confidence fluctuates significantly, just beyond the model's initial robustness and before saturation effects dominate in gradient-based methods. This choice ensures that \( D_{\alpha} \) targets a meaningful and interpretable point of model behavior. The choice is further supported by empirical ablation: we sweep \( \alpha \in \{0.3, 0.5, 0.7\} \) on all seven backbones and find WG to be robust across this range (\ref{appendix:alpha_ablation}).

\textbf{Optimization:} 
For computational efficiency, we employ above binary search algorithm to compute \( D_{\alpha} \), achieving a time complexity of \( O(\log n) \), where \( n \) is the number of features. We set a convergence threshold of \( \epsilon = 0.005 \) and limit the number of iterations to 100, ensuring both precision and scalability.  

\begin{figure}[tp]
    \centering
        \includegraphics[width=\linewidth]{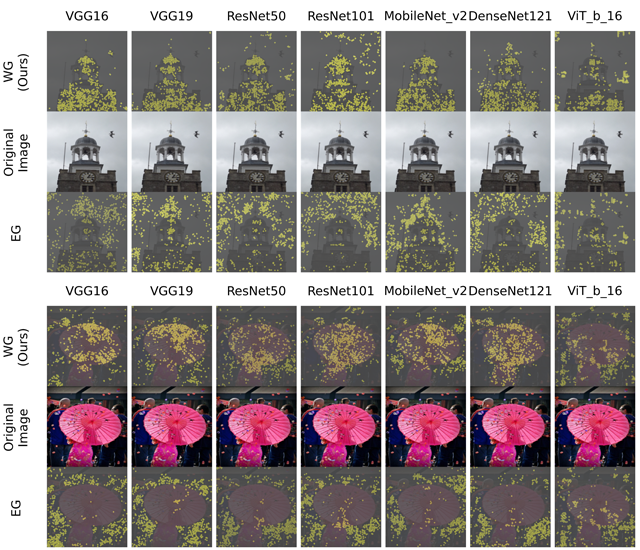}
    \caption{Saliency map comparisons across different models (columns) and methods (rows). The top row is our proposed method, while the bottom row employs Expected Gradients. Our proposed method, Weighted Integrated Gradients (WG), produces maps that focus more sharply on the main object than Expected Gradients in these examples, supporting improved attribution quality across the evaluated models.}
    \label{fig:ten_label}
\end{figure}

\subsection{Qualitative Analysis}

\textbf{a. Evaluating on AUC of Deletion and Overlap score}
Table~\ref{tb:auccompare} compares Expected Gradients (EG) and our proposed Weighted Integrated Gradients (WG) across seven backbone models using two metrics: AUC of Deletion (lower is better) and AUC of Overlap (higher is better). Reported values include mean $\pm$ standard deviations, relative improvements (Rel.I), and $p$-values from significance t-tests.

\rev{WG obtains lower Deletion AUC than EG on all evaluated models, with paired tests indicating statistically significant differences under this protocol.} For instance, on DenseNet121, WG reduces the AUC from 0.084 to 0.064 (24.03\% improvement, $p = 0.000$). Similar improvements are observed on MobileNetV2 (29.57\%, $p = 0.011$), ResNet50 (34.35\%, $p = 0.002$), ResNet101 (36.01\%, $p = 0.000$), VGG16 (23.25\%, $p = 0.002$), VGG19 (24.63\%, $p = 0.018$), and ViT-B/16 (19.36\%, $p = 0.024$), indicating WG’s improved ability to identify impactful features.

\rev{WG also obtains higher Overlap AUC than EG across the evaluated models, indicating stronger spatial agreement with the available segmentation masks under the selected protocol.} The largest gains are on ResNet101 (18.55\%), VGG16 (15.33\%), and ResNet50 (14.85\%), with over 10\% improvement on MobileNetV2 (10.82\%) and DenseNet121 (10.59\%); smaller but still significant gains are observed on VGG19 (7.38\%) and ViT-B/16 (5.69\%), all statistically significant.

\rev{Further analysis shows consistent WG gains}: on shallow architectures like VGG16/19, it brings modest gains in Overlap AUC but significant improvements in Deletion AUC, reflecting better feature selection. On deeper models such as DenseNet121 and ResNet101, WG achieves larger gains on both metrics, demonstrating its scalability and effectiveness in modeling complex feature hierarchies. For visualization, Fig. \ref{fig:ten_label} illustrates that our method, Weighted Integrated Gradients (WG), produces sharper and more relevant saliency maps compared to the baseline Expected Gradients across various models.

\rev{These results support the hypothesis that fitness-based baseline weighting can improve attribution metrics relative to uniform baseline averaging; however, the gains should be interpreted together with the additional computational cost of estimating the fitness scores and the limitations of the evaluation metrics.} WG achieves an average 27.31\% improvement in Deletion AUC and 11.89\% in Overlap AUC across the seven backbones. \rev{WG requires estimating $D_\alpha$ for each baseline, which involves extra model evaluations beyond the IG computations used by EG; it is therefore most appropriate when attribution fidelity and baseline adaptivity are prioritized over minimal runtime rather than as a faster alternative to EG.}

\section{Conclusion}
In this study, we introduced Weighted Integrated Gradients (WG) for feature attribution. Unlike previous approaches that treat all baselines equally, WG autonomously evaluates the relevance of each baseline and assigns appropriate weights via an unsupervised fitness measure. WG meets essential requirements for a feature attribution method, including completeness, sensitivity, and implementation invariance, while also providing an assumption-dependent, proxy-based probabilistic justification for increasing expected relevance mass relative to uniform baseline averaging. \rev{Experimental results on metrics, such as AUC of Deletion and Overlap, showed WG gains over evaluated methods across convolutional and Transformer architectures, with consistent improvements when combined with complementary attribution methods under the adopted protocol. At the same time, the theory rests on an expected, proxy-based fitness--relevance monotonicity assumption and is not a per-instance guarantee; WG adds fitness-evaluation overhead and is noticeably slower than EG ($\sim\!5.7\times$ in our measurements), so it is best viewed as an attribution-fidelity trade-off rather than an efficient replacement; and Overlap AUC depends on ground-truth masks and correctly classified samples and should be read alongside Deletion AUC.} For future work, \rev{evaluating WG under broader faithfulness protocols—including ground-truth-free metrics, misclassified samples, distribution shifts, and latency-constrained settings—is a promising direction to pursue.}

\vspace{-10pt}
\section*{Acknowledgement}
This research is funded by Hanoi University of Science and Technology (HUST) under
project number T2024-PC-040.

\vspace{-10pt}
\section*{Declaration of generative AI and AI-assisted technologies in the manuscript preparation process.}

During the preparation of this work the author(s) used ChatGPT in order to copyediting and text refinement. After using this tool/service, the author(s) reviewed and edited the content as needed and take(s) full responsibility for the content of the published article.

\appendix

\vspace{-10pt}
\section{Proofs}
\label{appendix:proof}
\begin{proof}[Proof of Proposition~\ref{prop:wg_improves_relevance}]
Reindex both aggregations by rank rather than baseline index. Since $r \mapsto k(x,r)$ is a bijection for each $x$, we have $\sum_k w_k(x) Q_k(x) = \sum_r W_r(x) Q_r(x)$ where $W_r(x) := w_{k(x,r)}(x)$ and $Q_r(x) := Q_{k(x,r)}(x)$. By construction of WG, $W_r(x)$ is nonincreasing in $r$ pointwise (rank $r=1$ has the smallest $D_\alpha$, hence the largest weight after normalization), and $\sum_r W_r(x) = 1$. Hence $\bar{W}_r := \mathbb{E}_x[W_r(x)]$ is nonincreasing in $r$ with $\sum_r \bar{W}_r = 1$. The rank-marginal WG profile $p^{\mathrm{WG}^{\star}}$ satisfies $\mathbb{E}_x[q_{\mathrm{WG}^{\star}}(x)] = \sum_r \bar{W}_r \bar{Q}_r$, and the EG side reduces to $\mathbb{E}_x[q_{\mathrm{EG}}(x)] = \frac{1}{n}\sum_r \bar{Q}_r$. By Assumption~\ref{ass:fitness_relevance}, $(\bar{Q}_r)$ is also nonincreasing in $r$. Applying Chebyshev's sum inequality to the two similarly-ordered deterministic sequences $(\bar{W}_r)$ and $(\bar{Q}_r)$, and using $\sum_r \bar{W}_r = 1$, yields $\sum_r \bar{W}_r \bar{Q}_r \ge \frac{1}{n}\sum_r \bar{Q}_r$, the claimed bound. Strictness holds whenever $(\bar{Q}_r)$ is non-constant, since Chebyshev's inequality is strict for non-constant similarly-ordered sequences. Intuitively, WG aligns higher rank-averaged weights with higher rank-averaged relevance, whereas EG weights all ranks equally.
\end{proof}

\begin{proof}[Proof of Theorem~\ref{thm:wg_probabilistic_rate}]
Fix $x$ with $\delta(x) > 0$ and let $J_t^{\mathrm{WG}}$ be i.i.d.\ from $p^{\mathrm{WG}}$, with indicators $X_t^{\mathrm{WG}} = \mathbf{1}\{J_t^{\mathrm{WG}} \in R(x)\} \in [0,1]$, so $\hat{q}_{\mathrm{WG}}(x,m) = \frac{1}{m}\sum_t X_t^{\mathrm{WG}}$ has mean $q_{\mathrm{WG}}(x)$. By Hoeffding's inequality, $\Pr(\hat{q}_{\mathrm{WG}}(x,m) \le q_{\mathrm{WG}}(x) - \varepsilon) \le e^{-2m\varepsilon^2}$ for any $\varepsilon > 0$. Setting $\varepsilon = \delta(x) = q_{\mathrm{WG}}(x) - \mu_{\mathrm{EG}} > 0$, which is admissible by the hypothesis, yields $q_{\mathrm{WG}}(x) - \varepsilon = \mu_{\mathrm{EG}}$ and hence $\Pr(\hat{q}_{\mathrm{WG}}(x,m) \le \mu_{\mathrm{EG}}) \le e^{-2m\delta(x)^2}$. Rearranging, for any $\eta\in(0,1)$, $m \ge \frac{1}{2\delta(x)^2}\log(1/\eta) \Rightarrow \Pr(\hat{q}_{\mathrm{WG}}(x,m) > \mu_{\mathrm{EG}}) \ge 1-\eta$.
\end{proof}

\vspace{-10pt}
\section{Empirical verification of Assumption~\ref{ass:fitness_relevance}}
\label{appendix:assumption_verification}

We use the COCO segmentation mask $S(x)$ as a proxy for $R(x)$, giving relevance quality $Q_k(x) = \sum_{j \in S(x)} p_j^{(k)}$ with $p_j^{(k)} = [\mathrm{IG}_j(x, x^{(k)})]_+ / \sum_\ell [\mathrm{IG}_\ell(x, x^{(k)})]_+$. For each input $x$, we rank the $n=6$ baselines by $D_\alpha(x^{(k)})$ (rank~1 = best fitness) and estimate $\mathbb{E}[Q_k \mid r(k) = r] = \frac{1}{|\mathcal{X}|}\sum_{x \in \mathcal{X}} Q_{k(x,r)}(x)$, which Assumption~\ref{ass:fitness_relevance} predicts to be non-increasing in $r$.

\begin{figure}[h]
\centering
\includegraphics[width=0.85\linewidth]{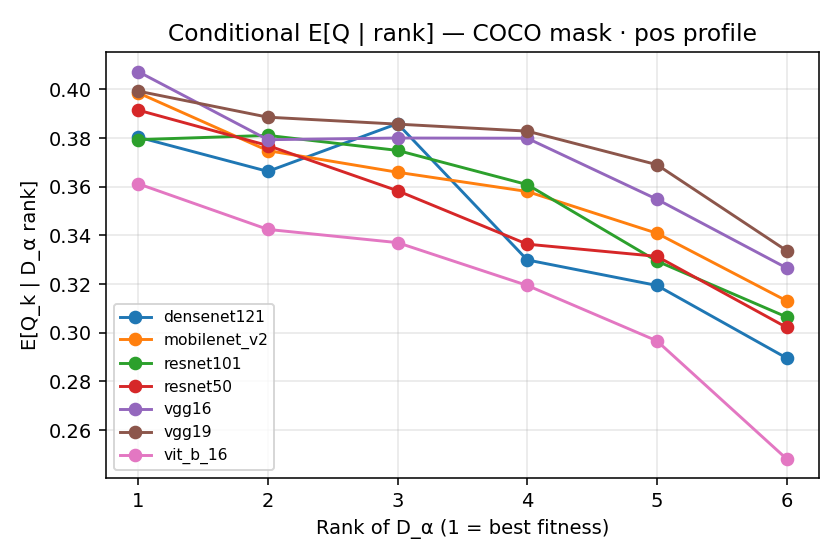}
\caption{Empirical validation across seven backbones: $\mathbb{E}[Q_k \mid r(k) = r]$ exhibits a decreasing trend on all backbones, including ViT-B/16.}
\label{fig:assumption_validation}
\end{figure}

Figure~\ref{fig:assumption_validation} confirms the predicted trend on all 7 backbones, strictly for ViT-B/16 and the ResNet family, with minor non-monotonicity at ranks~2--3 for VGG16/19. Local failures stem from \emph{saturated baselines} ($D_\alpha$ differences within binary-search noise) or \emph{spurious baselines} (low $D_\alpha$ from non-causal features), both of which average out across inputs, justifying the in-expectation form of Assumption~\ref{ass:fitness_relevance}. \rev{This proxy does not necessarily capture all causally relevant pixels; the validation should be read as empirical support for the expected proxy monotonicity assumption, not as proof of true causal relevance.}

\vspace{-10pt}
\section{Filtered-WG: a quality ablation}
\label{appendix:filtered_wg}

We assess whether the fitness score $D_\alpha$ correctly identifies disposable baselines by comparing three aggregation schemes: \text{EG} (uniform $1/n$ weighting), \text{WG} (fitness weighting $w_k \propto 1/D_\alpha$), and \text{Filtered-WG} (WG with low-fitness baselines zeroed out before aggregation, i.e.\ $w_k < 0.25\,\overline{w}$).

\paragraph{Quality results}
Table~\ref{tab:filtered_wg_avg} reports the mean AUC of Deletion and Overlap across the seven backbones. WG substantially improves over EG, and Filtered-WG matches WG to within $\sim 5\times 10^{-4}$ (in fact marginally better) on both metrics, confirming that $D_\alpha$ correctly identifies non-informative baselines: dropping them does not harm and can sharpen the attribution.

\begin{table}[htbp]
\centering
\fontsize{9}{9}\selectfont
\begin{tabular}{lccc}
\hline
\textbf{Metric} & \textbf{EG} & \textbf{WG} & \textbf{Filtered-WG} \\
\hline
Deletion ($\downarrow$) & 0.0575 & 0.0428 & \textbf{0.0423} \\
Overlap ($\uparrow$)    & 0.4071 & 0.4554 & \textbf{0.4558} \\
\hline
\end{tabular}
\caption{Mean AUC of Deletion and Overlap averaged over the seven backbones. Filtered-WG matches WG to within $\sim 5\times 10^{-4}$ on both metrics.}
\label{tab:filtered_wg_avg}
\end{table}

\vspace{-10pt}
\paragraph{Remark on cost}
\rev{Filtered-WG is a quality ablation, not an efficiency gain: $D_\alpha$ and the IG map are still computed for every baseline before filtering, so only the negligible final aggregation is saved.} Averaged over seven backbones (NVIDIA L4, PyTorch 2.5.1 + CUDA 12.1), WG costs about $3.45$\,s per image versus $0.60$\,s for EG, roughly a $5.7\times$ wall-clock overhead. \rev{WG is therefore an attribution-fidelity trade-off, not a faster variant.}

\vspace{-10pt}
\section{Comparison with broader XAI baselines}
\label{appendix:broader_baselines}

We compare WG against two prominent XAI methods: \text{SmoothGrad} (noise-averaging) \cite{smilkov2017smoothgrad} and \text{IDGI} \cite{yang2023idgi}(path step-weighting). Since WG (baseline weighting) is orthogonal to both, we also construct hybrid variants. All experiments use the same seven backbones, $n=6$ baselines, and AUC of Deletion / Overlap metrics as the main paper.

\paragraph{SmoothGrad}
SmoothGrad averages gradients over $n_{\mathrm{noise}}$ noisy copies $x+\eta_i$ with $\eta_i \sim \mathcal{N}(0,\sigma^2 I)$. We define WG-smooth by averaging IG over noisy inputs across fitness-weighted baselines, $\text{WG-smooth}_j(x) = \sum_{k=1}^{K} \frac{w_k}{n_{\mathrm{noise}}} \sum_{i=1}^{n_{\mathrm{noise}}} \mathrm{IG}_j(x + \eta_i;\, x'^{(k)})$, with $w_k \propto 1/D_\alpha(x'^{(k)})$, $n_{\mathrm{noise}}=8$, $\sigma = 0.15(x_{\max}-x_{\min})$. {EG-smooth} replaces $w_k$ with $1/K$ for isolation. Table~\ref{tab:sg_idgi_combined} reports pooled means. WG-smooth strictly dominates SmoothGrad on \emph{both} metrics across all 7 backbones; EG-smooth underperforms SmoothGrad on Overlap for 4/7 backbones, whereas WG-smooth wins on 7/7, with the largest gain on the Transformer backbone ViT-B/16.

\paragraph{IDGI}
IDGI re-weights gradients along the IG path by per-step contribution $\Delta f_k/\|g_k\|^2$ and is single-baseline in its original form. We construct \emph{WG+IDGI} by aggregating fitness-weighted IDGI maps across baselines, $\text{WG+IDGI}_j(x) = \sum_{k=1}^{K} w_k \cdot \mathrm{IDGI}_j(x; x'^{(k)})$ with $w_k \propto 1/D_\alpha(x'^{(k)})$, using $N_{\mathrm{IDGI}}=50$ straight-line steps as in Yang et al.\ (2023). \emph{EG+IDGI} uses $1/K$ weights. Table~\ref{tab:sg_idgi_combined} reports pooled results. Multi-baseline aggregation alone (EG+IDGI vs.\ single-baseline IDGI) already improves both metrics, showing IDGI is itself baseline-sensitive; adding fitness weighting (WG+IDGI) gives a further gain, for a total of about $27.9\%$ Deletion improvement over single-baseline IDGI (significant on 7/7 backbones).

\begin{table}[htbp]
\centering
\fontsize{8}{9}\selectfont
\begin{tabular}{llccc}
\hline
\textbf{Family} & \textbf{Metric} & \textbf{Base} & \textbf{EG variant} & \textbf{WG variant} \\
\hline
\multirow{2}{*}{SmoothGrad}
  & Overlap ($\uparrow$)   & 0.4820 & 0.4850 & \textbf{0.5190} \\
  & Deletion ($\downarrow$) & 0.0799 & 0.0241 & \textbf{0.0183} \\
\hline
\multirow{2}{*}{IDGI}
  & Overlap ($\uparrow$)   & 0.5323 & 0.6121 & \textbf{0.6198} \\
  & Deletion ($\downarrow$) & 0.0695 & 0.0526 & \textbf{0.0501} \\
\hline
\end{tabular}
\caption{Pooled means across seven backbones for SmoothGrad and IDGI variants. \emph{Base} is the original single-baseline method; \emph{EG variant} adds uniform multi-baseline aggregation; \emph{WG variant} adds fitness-weighted multi-baseline aggregation. All improvements significant at $p-{value} < 0.05$.}
\label{tab:sg_idgi_combined}
\end{table}

\vspace{-10pt}
\section{Sensitivity to the fitness threshold $\alpha$}
\label{appendix:alpha_ablation}

To verify that WG is robust to the fitness threshold $\alpha$ (the target fraction of model output remaining after masking, set to $0.5$ following the toxicology LC50/LD50 convention), we sweep $\alpha \in \{0.3, 0.5, 0.7\}$ on all seven backbones with IG, baselines, and seed fixed, so that only $D_\alpha$, the weights $w_k$, and the downstream metrics vary.

\begin{table}[htbp]
\centering
\fontsize{8}{8}\selectfont
\begin{tabular}{lccccc}
\hline
\textbf{Model} & \textbf{EG} & $\mathbf{WG_{0.3}}$ & $\mathbf{WG_{0.5}}$ & $\mathbf{WG_{0.7}}$ & \textbf{Best $\alpha$} \\
\hline
\multicolumn{6}{l}{\textit{AUC of Deletion ($\downarrow$)}} \\
VGG16        & 0.035 & 0.027          & 0.027          & \textbf{0.026} & 0.7 \\
VGG19        & 0.029 & 0.022          & 0.022          & \textbf{0.022} & 0.7 \\
ResNet50     & 0.044 & \textbf{0.029} & 0.029          & 0.032          & 0.3 \\
ResNet101    & 0.057 & 0.039          & \textbf{0.037} & 0.037          & \textbf{0.5} \\
MobileNetV2  & 0.021 & 0.015          & \textbf{0.014} & 0.015          & \textbf{0.5} \\
DenseNet121  & 0.084 & 0.064          & 0.064          & \textbf{0.063} & 0.7 \\
ViT-B/16     & 0.132 & 0.115          & 0.106          & \textbf{0.105} & 0.7 \\
\hline
\multicolumn{6}{l}{\textit{AUC of Overlap ($\uparrow$)}} \\
VGG16        & 0.423 & 0.484          & \textbf{0.488} & 0.485          & \textbf{0.5} \\
VGG19        & 0.427 & \textbf{0.467} & 0.459          & 0.450          & 0.3 \\
ResNet50     & 0.390 & 0.444          & 0.448          & \textbf{0.453} & 0.7 \\
ResNet101    & 0.391 & 0.463          & \textbf{0.463} & 0.457          & \textbf{0.5} \\
MobileNetV2  & 0.414 & 0.456          & \textbf{0.459} & 0.455          & \textbf{0.5} \\
DenseNet121  & 0.418 & \textbf{0.463} & 0.463          & 0.462          & 0.3 \\
ViT-B/16     & 0.387 & 0.406          & \textbf{0.409} & 0.402          & \textbf{0.5} \\
\hline
\end{tabular}
\caption{AUC of Deletion and Overlap under varying $\alpha$. WG beats EG for every $\alpha \in \{0.3, 0.5, 0.7\}$ on every backbone.}
\label{tab:alpha_ablation}
\end{table}

The results are shown in Table \ref{tab:alpha_ablation}. For every $\alpha \in \{0.3, 0.5, 0.7\}$, WG beats EG on every backbone on both metrics. Among the three values, $\alpha = 0.5$ is the most frequent strict best across the (model, metric) cells and is the strict best for the Transformer backbone ViT-B/16 on Overlap, while differences within $\{0.3, 0.5, 0.7\}$ are small and rarely significant after Bonferroni correction. We therefore retain $\alpha = 0.5$ as the default: it is the most frequent empirical optimum and consistent with the toxicology motivation, so a practitioner can safely pick it without dataset-specific tuning.
\vspace{-10pt}
\section{Extension to object detection}
\label{appendix:detection}
WG extends to object detection by redefining only the scalar target $f$ that IG attributes to; all other components (Algorithm~1 binary search, baseline library $D'$, weight normalization) are unchanged. We use the pre-softmax class logit of the top-1 detection, $f_{\det}(x) = \mathrm{logits}[i^\star, c^\star]$ with $(i^\star, c^\star) = \arg\max_{i,c} \sigma(\mathrm{logits}_{i,c}(x))$, where $i$ indexes the detection slot (anchor for RetinaNet, query for RT-DETR) and $c$ the class. The fitness $D_\alpha$ and AUC of Deletion both use $\sigma(\mathrm{logits}) \in [0,1]$ to keep the deletion curve in the same range as the classifier setting.
\vspace{-10pt}
\paragraph{Setup}
We evaluate on two architecturally distinct detectors: a CNN single-stage anchor-based model (\text{RetinaNet-R50-FPN}, torchvision) and a Transformer encoder--decoder model (\text{RT-DETR-R50}, PekingU/rtdetr\_r50vd). The evaluation uses 80 COCO images at $640 \times 640$ with segmentation masks, $M = 20$ Riemann samples, and the same six baselines as the classification experiments. Predictions are filtered by top-1 confidence $\ge 0.3$ and $\mathrm{IoU}(\text{pred}, \text{GT}) \ge 0.1$.

\begin{table}[htbp]
\centering
\fontsize{8}{8}\selectfont
\begin{tabular}{llccc}
\hline
\textbf{Detector} & \textbf{Metric} & \textbf{EG} & \textbf{WG (Ours)} & \textbf{Rel. I (\%)} \\
\hline
\multirow{2}{*}{RetinaNet-R50-FPN}
  & Deletion ($\downarrow$) & 0.0941 & \textbf{0.0782} & 16.9 \\
  & Overlap ($\uparrow$)    & 0.3889 & \textbf{0.3983} & 2.4 \\
\hline
\multirow{2}{*}{RT-DETR-R50}
  & Deletion ($\downarrow$) & 0.0142 & \textbf{0.0127} & 10.5 \\
  & Overlap ($\uparrow$)    & 0.3659 & \textbf{0.4249} & 16.1 \\
\hline
\end{tabular}
\caption{WG vs.\ EG on object detection. For Deletion, smaller is better; for Overlap, higher is better.}
\label{tab:detection}
\end{table}

The results are shown in Table \ref{tab:detection}. On both detectors, WG outperforms EG in the same direction as in classification, with relative improvements in roughly the $10$--$17\%$ range—squarely within the classification range. This confirms that WG generalises naturally to dense prediction tasks: only the scalar target $f$ changes, while the rest of the framework is unmodified.

{\fontsize{9}{9} \selectfont
\bibliographystyle{elsarticle-num} 
\bibliography{ref.bib}

@inproceedings{Sundararajan16,
  title={Axiomatic attribution for deep networks},
  author={Sundararajan, Mukund and Taly, Ankur and Yan, Qiqi},
  booktitle={International conference on machine learning},
  pages={3319--3328},
  year={2017},
  organization={PMLR}
}

@article{Erion21,
  title={Improving performance of deep learning models with axiomatic attribution priors and expected gradients},
  author={Erion, Gabriel and Janizek, Joseph D and Sturmfels, Pascal and Lundberg, Scott M and Lee, Su-In},
  journal={Nature machine intelligence},
  volume={3},
  number={7},
  pages={620--631},
  year={2021},
  publisher={Nature Publishing Group UK London}
}

@article{Li16,
  title   = {Perturbation-Based Methods for Explaining Deep Neural Networks: A Survey},
  author  = {Ivanovs, Maksims and Kadikis, Roberts and Ozols, Kaspars},
  journal = {Pattern Recognition Letters},
  volume  = {150},
  pages   = {228--234},
  year    = {2021},
  publisher = {Elsevier}
}

@inproceedings{Vig19,
  title={BertViz: A tool for visualizing multihead self-attention in the BERT model},
  author={Vig, Jesse},
  booktitle={ICLR workshop: Debugging machine learning models},
  volume={3},
  year={2019}
}

@inproceedings{Akhtar23,
  title={Towards credible visual model interpretation with path attribution},
  author={Akhtar, Naveed and Jalwana, Mohammad AAK},
  booktitle={International Conference on Machine Learning},
  pages={439--457},
  year={2023},
  organization={PMLR}
}

@inproceedings{Xu20,
  title={Attribution in scale and space},
  author={Xu, Shawn and Venugopalan, Subhashini and Sundararajan, Mukund},
  booktitle={Proceedings of the IEEE/CVF Conference on Computer Vision and Pattern Recognition},
  pages={9680--9689},
  year={2020}
}

@article{klein2024navigating,
  title={Navigating the maze of explainable ai: A systematic approach to evaluating methods and metrics},
  author={Klein, Lukas and L{\"u}th, Carsten and Schlegel, Udo and Bungert, Till and El-Assady, Mennatallah and J{\"a}ger, Paul},
  journal={Advances in Neural Information Processing Systems},
  volume={37},
  pages={67106--67146},
  year={2024}
}

@article{dwivedi2024efficient,
  title={An efficient ensemble explainable AI (XAI) approach for morphed face detection},
  author={Dwivedi, Rudresh and Kothari, Pranay and Chopra, Deepak and Singh, Manjot and Kumar, Ritesh},
  journal={Pattern Recognition Letters},
  volume={184},
  pages={197--204},
  year={2024},
  publisher={Elsevier}
}

@inproceedings{sundararajan2020many,
  title={The many Shapley values for model explanation},
  author={Sundararajan, Mukund and Najmi, Amir},
  booktitle={International conference on machine learning},
  pages={9269--9278},
  year={2020},
  organization={PMLR}
}

@inproceedings{koenen2024toward,
  title={Toward understanding the disagreement problem in neural network feature attribution},
  author={Koenen, Niklas and Wright, Marvin N},
  booktitle={World Conference on Explainable Artificial Intelligence},
  pages={247--269},
  year={2024},
  organization={Springer}
}

@article{sturmfels2020visualizing,
  title={Visualizing the impact of feature attribution baselines},
  author={Sturmfels, Pascal and Lundberg, Scott and Lee, Su-In},
  journal={Distill},
  volume={5},
  number={1},
  pages={e22},
  year={2020}
}

@article{wang2026twin,
  title={A twin-branch decoupled network for multi-class unsupervised anomaly detection},
  author={Wang, Bohan and Wan, Jihong and Zhao, Jie and Ouyang, Xiaocao and Li, Xiaoping},
  journal={Engineering Applications of Artificial Intelligence},
  volume={167},
  pages={113891},
  year={2026},
  publisher={Elsevier}
}

@article{wan2026uncertainty,
  title={Uncertainty-aware significance and interaction-enhanced feature selection: A fuzzy multi-granularity information perspective},
  author={Wan, Jihong and Li, Xiaoping and Wang, Zhihong and Zhao, Jie},
  journal={Information Processing \& Management},
  volume={63},
  number={6},
  pages={104775},
  year={2026},
  publisher={Elsevier}
}

@article{smilkov2017smoothgrad,
  title={Smoothgrad: removing noise by adding noise},
  author={Smilkov, Daniel and Thorat, Nikhil and Kim, Been and Vi{\'e}gas, Fernanda and Wattenberg, Martin},
  journal={arXiv preprint arXiv:1706.03825},
  year={2017}
}

@inproceedings{yang2023idgi,
  title={Idgi: A framework to eliminate explanation noise from integrated gradients},
  author={Yang, Ruo and Wang, Binghui and Bilgic, Mustafa},
  booktitle={Proceedings of the IEEE/CVF Conference on Computer Vision and Pattern Recognition},
  pages={23725--23734},
  year={2023}
}
}

\end{document}